\documentclass[a4paper,fleqn]{cas-sc}

\usepackage[utf8]{inputenc}
\usepackage[T1]{fontenc}
\usepackage[numbers]{natbib}
\usepackage{tikz}
\usetikzlibrary{arrows.meta}
\usepackage{placeins}

\begin{document}
\let\WriteBookmarks\relax
\def\floatpagepagefraction{1}
\def\textpagefraction{.001}

\shorttitle{Open-vocabulary 3D object detection with promptable segmentation}
\shortauthors{Ö.F. Deniz and M.T. Koçyiğit}
\let\printorcid\relax

\title[mode = title]{Open-vocabulary 3D object detection with promptable
segmentation}

\author[1]{{Ö}mer Faruk Deniz}[type=author]
\ead{omer.deniz@std.bogazici.edu.tr}
\credit{Conceptualization, Methodology, Software, Investigation, Writing -- original draft}

\author[1]{Mustafa Taha Koçyiğit}
\cormark[1]
\ead{taha.kocyigit@bogazici.edu.tr}
\credit{Supervision, Writing -- review and editing}

\affiliation[1]{organization={Institute of Data Science and Artificial
Intelligence, Boğaziçi University},
addressline={South Campus, Bebek},
postcode={34342},
city={Istanbul},
country={Türkiye}}

\cortext[1]{Corresponding author.}

\begin{abstract}
Three-dimensional object detection for autonomous driving is dominated by
detectors trained on large corpora of human-annotated 3D boxes. Such a
detector learns a fixed category list, and everything outside it is
invisible. This paper asks whether the task can be solved training-free
and open-vocabulary. A promptable segmentation model (SAM3), queried with
class names as text prompts, supplies instance masks in the vehicle's six
surround-view cameras, and the masks are turned into metric 3D boxes using
the geometry of the scene. The core is a controlled three-stage comparison
on nuScenes in which 2D detection is held fixed and only the source of 3D
geometry changes. Geometry predicted from images alone reaches 0.183 mean
average precision (mAP) under the official protocol; fitting boxes from raw
LiDAR points inside the same masks with training-free rules reaches
0.298~mAP / 0.348 nuScenes detection score (NDS) at zero labeling cost;
borrowing supervised box
geometry at inference time lifts the same detections to 0.413~mAP /
0.555~NDS, which locates the pipeline's largest deficit in measurement
precision rather than 2D detection, while class confusion and confidence
calibration survive that substitution. Reversing the direction, a
three-state camera-witness rule built from the same masks improves a
supervised LiDAR-only detector from 0.596 to 0.630~mAP, roughly half the
gain of fully supervised camera fusion, with no training. A coverage
analysis shows that SAM3 finds 84\% of in-range objects with a correctly
named mask; the classes that fail in the official metric are misnamed or
geometrically unforgiving, not unseen.
\end{abstract}


\maketitle

\section{Introduction}
\label{sec:introduction}

An autonomous vehicle acts on what it perceives. Every driving decision
ultimately rests on a list of the objects around the vehicle, each
carrying a position, a size, an orientation and a velocity, all measured
in meters relative to the vehicle. Producing this list from on-board
sensors is the task of three-dimensional (3D) object detection, and an
error at this stage propagates directly into behavior: an undetected
pedestrian is a pedestrian the planner will not brake for.

The dominant way to build such a detector is supervised learning: deep
networks trained on large corpora of human-annotated 3D
boxes~\citep{pointpillars,centerpoint}. This approach works remarkably well,
but it carries two structural costs. The first is annotation. Labeling
sequences from a LiDAR (Light Detection and Ranging) sensor with oriented 3D
boxes is among the most expensive labeling tasks in computer vision, and
the effort does not transfer: a new sensor arrangement, a new city or a
revised class list restarts it.

The second cost is subtler: a \emph{closed vocabulary}. A detector trained
on ten classes does not merely perform worse on an eleventh; it is
structurally incapable of reporting it. Traffic, however, is not limited to
ten categories: ambulances and fire trucks behave unlike ordinary trucks, a
child is not an average pedestrian, and road debris belongs to no class at
all. The class list of the benchmark used in this paper,
nuScenes~\citep{nuscenes}, excludes police vehicles, so a system that
places a visually correct box on one is \emph{penalized} for that
detection. Extending a closed-vocabulary detector to any of these concepts
means new labels and retraining.

Meanwhile, image understanding has moved in the opposite direction.
Foundation models trained on web-scale image--text data recognize concepts
that are named in free-form text rather than fixed in advance, an ability
referred to as an \emph{open vocabulary}. Promptable segmentation models
such as SAM3~\citep{sam3} take this one step further: given a text prompt,
they return a mask for every instance of the named concept in the image.
These models, however, return pixels, not positions, while driving
requires metric 3D. The missing element is not geometry itself, since the
vehicle already measures its surroundings with LiDAR, but the link between
a named concept in the image and a metric box in the world.

This paper asks how far that gap can be closed without any 3D annotation.
The system it studies uses only text prompts, a promptable segmenter, and
raw sensor geometry. Can such a system produce useful 3D detections, and
what exactly separates it from supervised detectors?

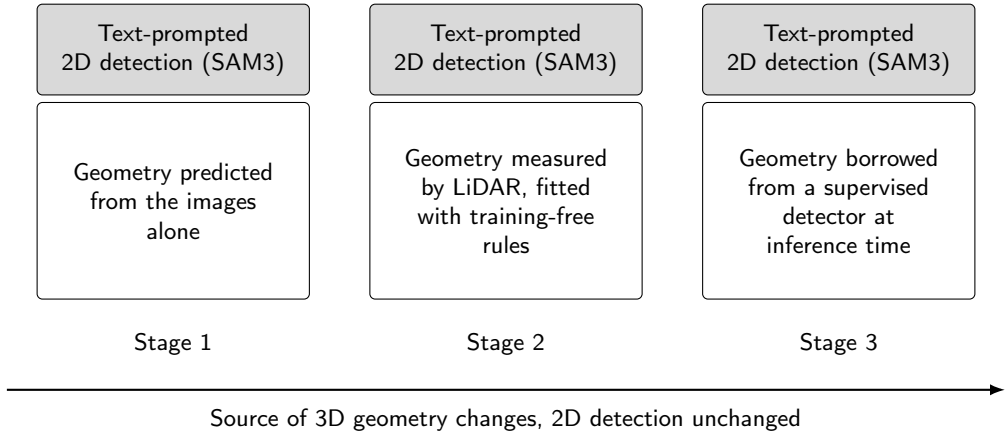
\begin{figure}
\centering
\begin{tikzpicture}[
  font=\small,
  rec/.style={draw, fill=black!15, rounded corners=2pt, align=center,
              inner sep=4pt, minimum height=12mm, minimum width=36mm},
  geo/.style={draw, rounded corners=2pt, align=center,
              inner sep=4pt, minimum height=26mm, minimum width=36mm},
  ar/.style={-{Latex[length=2mm]}, thick}
]
\node[rec] (r1) at (0,0)   {Text-prompted\\2D detection (SAM3)};
\node[rec] (r2) at (4.4,0) {Text-prompted\\2D detection (SAM3)};
\node[rec] (r3) at (8.8,0) {Text-prompted\\2D detection (SAM3)};
\node[geo] (g1) at (0,-2.0)   {Geometry predicted\\from the images\\alone};
\node[geo] (g2) at (4.4,-2.0) {Geometry measured\\by LiDAR, fitted\\with training-free\\rules};
\node[geo] (g3) at (8.8,-2.0) {Geometry borrowed\\from a supervised\\detector at\\inference time};
\node at (0,-3.9)   {Stage 1};
\node at (4.4,-3.9) {Stage 2};
\node at (8.8,-3.9) {Stage 3};
\draw[ar] (-2.2,-4.5) -- (11.0,-4.5);
\node[font=\small] at (4.4,-4.9) {Source of 3D geometry changes, 2D detection unchanged};
\end{tikzpicture}
\caption{The three-stage comparison. The 2D detection is the same in all
three stages, and only the source of the 3D geometry changes. The shaded box
marks the component that is held fixed.}
\label{fig:ladder}
\end{figure}

The central idea is a three-stage comparison. Each stage is a complete
version of the pipeline, evaluated on the nuScenes benchmark under its
official protocol. Text-prompted 2D detection is held fixed across the
stages, and only the source of the metric 3D geometry changes: in the
first stage it is predicted from the images alone, in the second it is
measured by LiDAR and fitted with training-free rules, and in the third it
is borrowed at inference time from a supervised detector. Because a single
factor changes at a time, the difference between two consecutive stages
isolates the value of that factor, which separates the value of range
sensing, of geometric fitting and of supervised geometry from the value of
open-vocabulary detection. Figure~\ref{fig:ladder} shows the three stages
side by side; the shaded box is identical in every stage, and only the
geometry below it changes.

In the final step the direction is reversed. The same masks are attached to
a supervised LiDAR-only detector as a training-free \emph{camera witness}:
for every box the detector reports, the system checks whether the cameras
contain a mask of the same class at the same place, and the confidence of
the box is kept or lowered accordingly. This configuration is compared
against a detector that learns its camera fusion from full supervision.

The main contributions of this paper are the following.
\begin{enumerate}
  \item A training-free, open-vocabulary 3D detection pipeline that turns
  promptable segmentation masks into 3D boxes with simple, class-aware rules
  applied to the LiDAR points, evaluated end-to-end under the official
  nuScenes protocol and accompanied by a study of prompt design and by an
  inference procedure that makes evaluation on the full validation split
  possible on a single GPU.
  \item A camera-only variant of the same pipeline, which separates the
  contribution of measured range from that of open-vocabulary detection.
  \item A method for attaching supervised box geometry to the pipeline's
  own detections at inference time, which shows that the largest deficit
  of the pipeline is measurement precision rather than 2D detection, while
  two further limits, class confusion and confidence calibration, survive
  that substitution.
  \item A training-free camera witness that improves a supervised
  LiDAR-only detector, recovering roughly half of the gain of fully
  supervised camera fusion at zero annotation cost.
  \item A coverage analysis of open-vocabulary 3D detection: which objects
  are found, which are misnamed, and where coverage ends.
\end{enumerate}

\section{Related work}
\label{sec:related}

\subsection{Supervised 3D object detection}

A LiDAR-based 3D detector takes as input a point cloud, the set of 3D
points measured by the spinning laser scanner, and outputs a list of
oriented 3D boxes, each with a class label and a confidence score. Modern
benchmarks---KITTI~\citep{kitti} and, at larger scale,
nuScenes~\citep{nuscenes}---established the now-standard formulation of
this task: oriented boxes over a fixed taxonomy, learned from dense human
annotation. VoxelNet~\citep{voxelnet} introduced end-to-end learning on
voxelized point clouds, SECOND~\citep{second} made the voxel backbone
practical with sparse convolutions, and pillar-based
encoders~\citep{pointpillars} made bird's-eye-view (top-down) processing
of the point cloud efficient. Center-based designs such as
CenterPoint~\citep{centerpoint} became the dominant paradigm on nuScenes:
objects are detected as center points in a bird's-eye-view map, and box
dimensions, orientation and velocity are read off from the features at
each center. CenterPoint plays two roles in this paper: it is the source
of supervised geometry in the third stage, and the LiDAR-only detector to
which the camera witness is attached.

Supervised detection also has a camera-only line. FCOS3D~\citep{fcos3d}
predicts 3D boxes from a single image with a fully convolutional head,
while DETR3D~\citep{detr3d} and BEVFormer~\citep{bevformer} aggregate
multi-camera evidence through learned queries and bird's-eye-view
representations. These detectors use the same input as the camera-only
variant studied here, but learn a closed vocabulary from full 3D
supervision, whereas the variant here is training-free and
open-vocabulary.

Multi-modal extensions inject camera information into the LiDAR pipeline.
PointPainting~\citep{pointpainting} decorates each LiDAR point with semantic
scores predicted from the images. MVP~\citep{mvp} densifies the point cloud
with camera-derived \emph{virtual points}, a design decision that matters
for this paper: it means that even ``LiDAR'' detectors internally exploit
dense image evidence. BEVFusion~\citep{bevfusion} fuses the two modalities
in a shared bird's-eye-view representation. All of these systems share the
two costs that motivate this study: full 3D supervision and a closed
vocabulary. The final part of this paper asks what a camera channel is
worth to a LiDAR detector from the opposite end of the supervision
spectrum, attaching the camera evidence at inference time through
open-vocabulary masks and a fixed three-state rule, with no fusion
training at all.

\subsection{The nuScenes benchmark and evaluation protocol}
\label{sec:related-nuscenes}

Collected in Boston and Singapore, nuScenes~\citep{nuscenes} is a
large-scale autonomous driving dataset. It contains 1{,}000 driving scenes
of 20 seconds each, recorded with six cameras covering the full horizon, a
32-beam LiDAR spinning at 20\,Hz, and five radars. Objects are annotated
with oriented 3D boxes over ten classes on \emph{keyframes}, the two
frames per second at which all sensors are synchronized and labeled, and
every object carries a visibility level recording how much of it is seen
from the cameras. The validation
split used throughout this paper contains 150 scenes with 6{,}019 keyframes
and roughly 122{,}000 in-range annotated objects.

Under the official protocol, a predicted box matches a ground-truth object
of the same class if their centers are closer than a threshold on the
ground plane. Average precision (AP) is computed at four thresholds, 0.5,
1, 2 and 4\,m, and averaged over thresholds and classes into mean average
precision (mAP). Matching by center distance rather than by box overlap
decouples \emph{finding} an object from estimating its exact extent, which
sparse LiDAR often cannot support. The quality of the matched boxes is
measured separately by five true-positive errors: the average translation
error (ATE, in meters), the average scale error (ASE), the average
orientation error (AOE), the average velocity error (AVE) and the average
attribute error (AAE). The nuScenes Detection Score (NDS) combines
everything with mAP carrying half the weight,
\begin{equation}
\mathrm{NDS} \;=\; \tfrac{1}{10}\Big[\,5\,\mathrm{mAP} \;+\;
\textstyle\sum_{\mathrm{mTP}}\big(1-\min(1,\mathrm{mTP})\big)\Big],
\end{equation}
so that a system is rewarded both for finding objects and for measuring
them precisely. Each class is evaluated only within a class-specific range:
50\,m for vehicles, 40\,m for pedestrians and two-wheelers, and 30\,m for
traffic cones and barriers. Annotated objects with no LiDAR or radar return
are excluded from evaluation.

\subsection{Promptable segmentation and open-vocabulary perception}

CLIP~\citep{clip} demonstrated that image--text pretraining yields
transferable open-vocabulary recognition. One line of work brought this
ability to 2D detection. ViLD~\citep{vild} distills CLIP knowledge into a
detector, Detic~\citep{detic} widens the vocabulary with image-level
labels, and GLIP~\citep{glip} and Grounding DINO~\citep{groundingdino}
treat detection as the grounding of free-form text phrases. The Segment
Anything family took the complementary segmentation-and-prompt side. The
original Segment Anything Model (SAM)~\citep{sam} introduced
class-agnostic segmentation from geometric prompts: a click or a box
yields a mask for whatever object lies there, without naming its
category. SAM2~\citep{sam2} extended the same interface to video.
SAM3~\citep{sam3} added the capability this paper depends on:
\emph{concept prompts}. Given a short noun phrase, the model returns
\emph{all} instances of that concept in the image, each with its own mask
and confidence score, which turns a segmentation model into an
open-vocabulary detector whose vocabulary is decided at inference time.
Its strength is that the class list becomes a list of words; its cost is
that inference scales with the number of queried concepts, since each
prompt is a separate query against the image. The pipeline developed here
uses SAM3 as its sole 2D detection component: all category knowledge
enters the system as text prompts, and everything downstream of the mask
is training-free geometry.

\subsection{Open-vocabulary and weakly supervised 3D detection}

A growing body of work transfers 2D vision--language knowledge to 3D, along
two main routes. The first \emph{distills}: image--language features are
projected into the point cloud and a 3D network is trained to reproduce
them, as in OpenScene~\citep{openscene} and CLIP2Scene~\citep{clip2scene}.
The second generates \emph{pseudo-labels}: 2D detections are lifted to 3D
boxes and used as training targets for a 3D detector, as in
OV-3DET~\citep{ov3det}. Both
routes still train a 3D network on the target domain, although the
pseudo-label generation step is close in spirit to the pipeline studied
here. This paper differs in two respects: the entire system is
inference-time only, with no 3D network trained at any stage, and it
provides a complete, official-protocol evaluation together with a
controlled decomposition that locates precisely what separates such
training-free systems from supervised ones.

\subsection{Camera-only metric geometry}

Estimating metric geometry from images alone is a long-standing problem.
Classical multi-view stereo requires overlapping viewpoints, and learned
monocular depth was for years tied to the dataset it was trained on, with
unreliable absolute scale. MiDaS~\citep{midas} showed that training across
mixed datasets yields robust relative depth, and DUSt3R~\citep{dust3r}
established the \emph{point-map} paradigm, a metric 3D coordinate for
every pixel predicted directly from images, on which recent visual
geometry transformers build. The Driving Visual Geometry
Transformer~\citep{dvgt2} produces metric point maps from surround-view
driving footage. It serves as the camera-only geometry backbone of the
first stage: it replaces the LiDAR cloud under an otherwise identical
pipeline, which isolates the contribution of true range sensing.

\section{Method}
\label{sec:method}

The method is organized as a three-stage comparison. The 2D detection
component is fixed throughout: an open-vocabulary, promptable segmentation
model (SAM3) that localizes text-described concepts in the vehicle's six
surround-view cameras (Section~\ref{sec:method-prompts}). What changes
from stage to stage is where the metric 3D geometry comes from: learned
monocular geometry only (Section~\ref{sec:method-dvgt}), raw LiDAR returns
with training-free geometric fitting (Section~\ref{sec:method-lidar}), or
the box geometry of a supervised LiDAR detector, associated to the
pipeline's detections at inference time
(Section~\ref{sec:method-cpbridge}). Finally, the direction is reversed:
the same open-vocabulary machinery is attached to the supervised detector
as a training-free \emph{camera witness}
(Section~\ref{sec:method-witness}). No component at any stage is trained
or fine-tuned on nuScenes; the only domain-specific inputs are the class
names and the sensor calibration and poses that the dataset provides.
Section~\ref{sec:method-params} collects the numeric values behind the
design choices and states where each comes from.

\subsection{Promptable segmentation and prompt design}
\label{sec:method-prompts}

\subsubsection{SAM3 as the 2D detection component}

SAM3~\citep{sam3} performs \emph{promptable concept segmentation}: given a
free-text prompt, it returns an exhaustive set of instance masks for that
concept, each with a confidence score, without any task-specific training.
This single capability supplies everything the pipeline needs from the
images: knowledge of the category, separation of one object from another,
and localization in 2D.

\begin{figure}
\centering
\includegraphics[width=\textwidth]{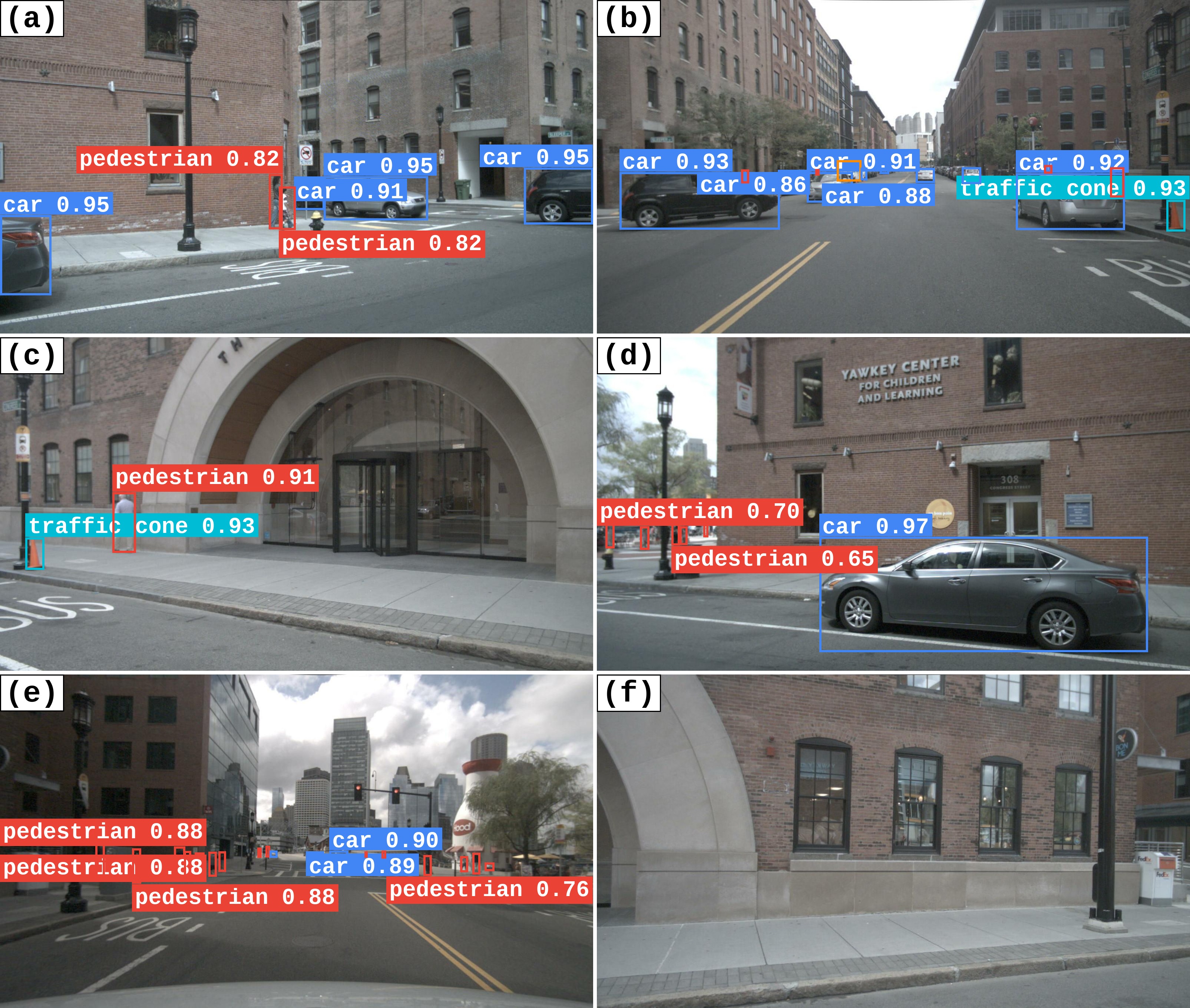}
\caption{SAM3 detections on the six surround-view cameras of a single
nuScenes keyframe~\citep{nuscenes}: (a) front left, (b) front, (c) front
right, (d) back left, (e) back, (f) back right. Each mask is drawn as its
bounding box, labeled with the prompt that produced it and with its score
where the label fits. Only detections scoring at least 0.5 are drawn, while
the pipeline itself uses every mask above the lower threshold.}
\label{fig:samsixcam}
\end{figure}

Figure~\ref{fig:samsixcam} shows SAM3 detections on the six cameras of one
validation frame. For readability the figure draws the bounding box of
each mask rather than the mask itself, labeled with the predicted class
and score; only detections scoring at least 0.5 are drawn, and repeated
detections of the same object are shown once, while the pipeline itself
uses every mask above the threshold described below.

A mask is not rejected on the strength of its score. Every mask that clears
the size and evidence tests of Section~\ref{sec:method-lidar} produces a
box, whatever its confidence, and the ranking is left to the evaluation.
The score of 0.2 in Table~\ref{tab:allparams} is a separate matter: it is
the confidence a detection must reach before the temporal layer of
Section~\ref{sec:method-postproc} uses it as evidence, as a node of a track
or as a neighbor from which a velocity is read. A detection below it is
still emitted and can still receive a velocity and an orientation from
more confident neighbors; it simply cannot anchor others.

Keeping the emission threshold low is worth measuring. On the full
validation split, lowering it from 0.5 to 0.2 roughly doubles the number
of boxes, from 143{,}012 to 277{,}291, and adds 0.031 mAP and 0.013 NDS.
The gain concentrates in the small classes, which the segmentation scores
least confidently: pedestrians gain 0.067 AP, motorcycles 0.063, traffic
cones 0.052 and bicycles 0.037, while cars gain 0.010 and trailers 0.001.
Section~\ref{sec:threshold-sensitivity} sweeps the threshold further. A
permissive list is safe because a low-scoring mask must still survive the
depth gate and the minimum-evidence test: masks that correspond to nothing
measurable are removed by the geometry stage rather than by the score.

\subsubsection{Prompt set}

Thirteen prompts are queried per image: one per nuScenes detection class,
plus three sub-class variants (Table~\ref{tab:prompts}). The variants
matter. The generic prompt ``construction vehicle'' is a weak description
of the excavators and bulldozers that dominate the class, and querying
those concepts directly yields measurably better detections; ``flatbed
trailer'' plays the same role for trailers. Detections produced by a
variant prompt are reported under their parent class: an ``excavator''
mask becomes a construction-vehicle detection.

\begin{table}[!htbp]
\centering
\caption{Text prompts used by the pipeline. The ten base prompts name the
evaluated classes. The three variant prompts describe subtypes and are
mapped onto the class they belong to.}
\label{tab:prompts}
\begin{tabular}{ll}
\hline
Role & Prompts \\
\hline
Base (10) & car, truck, bus, pedestrian, bicycle, motorcycle, \\
          & traffic cone, barrier, trailer, construction vehicle \\
Variants (3) & excavator, bulldozer, flatbed trailer \\
\hline
\end{tabular}
\end{table}

Prompt wording is not cosmetic. An early version of the pipeline used the
prompt ``person'', which turned out to segment \emph{depictions} of
people, such as figures on advertising billboards, alongside actual road
users; switching to ``pedestrian'' aligned the concept with the intended
traffic participant. The choice of prompt thus acts as a zero-cost prior
on what the system considers an object, and prompt engineering replaces
what would otherwise require retraining in a closed-vocabulary detector.

\subsubsection{Inference cost}

The flexibility of text prompts has a price: inference cost grows with the
number of queried concepts, since each prompt is a separate query against
the image. This cost is reduced in three ways: each image is encoded once
and all thirteen prompts run against that shared encoding, the six cameras
of a keyframe are processed as a single batch, and data preparation and
box fitting are overlapped with GPU inference. Together these measures
make the pipeline roughly fifteen times faster than the initial
implementation on equal hardware, so that a full run over the 150-scene
validation split completes in about one hour on a single GPU, with no
measurable change in accuracy.

\subsection{Camera-only lifting with learned monocular geometry}
\label{sec:method-dvgt}

The first stage uses no range sensor. \emph{Lifting} means turning a 2D
mask in the image into a 3D box in the world, and each stage is named after
the source of the 3D information it lifts with. The Driving Visual Geometry
Transformer (DVGT-2)~\citep{dvgt2} predicts a metric \emph{point map}---a
3D coordinate per pixel---from surround-view footage. Each SAM3 mask is
lifted by reading the predicted 3D points at its pixels and taking their
median as the object's center, and the box is completed with class-prior
dimensions, the typical size of the class. This variant is fully
camera-only: both the detections and the geometry come from images.
Everything downstream of the lifted center, namely the class-prior
dimensions, duplicate removal and the temporal layer, is shared with the
LiDAR-based variant, so the two stages differ only in where their geometry
comes from.

Its failure mode is instructive. Monocular geometry is \emph{inferred}
rather than measured, since metric scale has to be read from appearance
alone, so range errors grow with distance and with object size, and the
predicted surface can drift smoothly with no local cue that it has done
so. The consequence, quantified in Section~\ref{sec:experiments}, is that
camera-only lifting finds objects well but cannot place them precisely
enough for the strict center-distance thresholds of the 3D metric.

The two components meet on the image grid. SAM3 produces its masks at the
full camera resolution of 1600$\times$900, while the point map is defined
on the model's own 512$\times$288 grid, so a mask is downsampled to that
grid before the points under it are read. DVGT-2 returns the map in the ego
frame of the first keyframe of the scene, whose pose carries the lifted
centers into the global frame of the evaluation.

\subsection{LiDAR lifting with a depth gate}
\label{sec:method-lidar}

The second stage keeps the same masks and class names, and replaces
predicted geometry with measured geometry. The single-sweep LiDAR cloud of
the keyframe, the points from one full rotation of the scanner, is
projected into each camera, and the points that fall inside a mask are
assigned to that mask. Two details make this association reliable:

\begin{itemize}
  \item \textbf{Depth gate.} A 2D mask is depth-blind: a mask over a
  distant object also covers foreground points on occluders, and points
  leaking through mask boundaries land on the background. Assigned points
  whose range deviates from the in-mask median by more than a
  class-dependent tolerance (Table~\ref{tab:params}) are discarded.
  \item \textbf{Minimum evidence.} A mask must cover at least the
  class-dependent pixel count of Table~\ref{tab:params}, and at least
  three of its assigned points must survive the depth gate, before it
  produces a detection; this suppresses masks that see an object the LiDAR
  cannot confirm.
\end{itemize}

\begin{table}[!htbp]
\centering
\small
\setlength{\tabcolsep}{4pt}
\caption{Per-class pipeline parameters. The depth-gate tolerance scales
with the physical length of the class. The range limits reproduce the
class-specific ranges of the official evaluation protocol. Classes whose
center is taken from the point median use no box fit and therefore no fit
threshold.}
\label{tab:params}
\begin{tabular}{lcccccc}
\hline
Class & Min.\ mask (px) & Depth gate (m) & Range (m) & Center from & Fit thr.\ (m) & Prior $L\times W\times H$ (m) \\
\hline
Car           & 244 & 2.5 & 50 & box fit  & 2.4 & $4.62\times1.95\times1.73$ \\
Truck         & 244 & 4.0 & 50 & box fit  & 3.0 & $6.93\times2.51\times2.84$ \\
Bus           & 244 & 6.0 & 50 & box fit  & 4.5 & $11.00\times2.94\times3.47$ \\
Trailer       & 244 & 6.0 & 50 & box fit  & 4.0 & $12.29\times2.90\times3.87$ \\
Constr.\ veh. & 244 & 4.0 & 50 & box fit  & 3.0 & $6.37\times2.85\times3.19$ \\
Pedestrian    & 100 & 1.5 & 40 & median   & --- & $0.73\times0.67\times1.77$ \\
Motorcycle    & 100 & 1.5 & 40 & median   & --- & $2.11\times0.77\times1.47$ \\
Bicycle       & 100 & 1.5 & 40 & median   & --- & $1.70\times0.60\times1.28$ \\
Traffic cone  &  50 & 1.0 & 30 & median   & --- & $0.41\times0.41\times1.07$ \\
Barrier       & 100 & --- & 30 & segments & --- & $0.50\times2.53\times0.98$ \\
\hline
\end{tabular}
\end{table}

\begin{figure}
\centering
\begin{tikzpicture}[font=\small]
\draw[-{Latex[length=2mm]}] (0,0) -- (11.2,0);
\node at (10.5,-0.4) {range (m)};
\draw[-{Latex[length=2mm]}] (0,0) -- (0,3.4);
\node[rotate=90] at (-0.4,1.7) {in-mask point count};
\fill[black!10] (4.6,0) rectangle (6.6,2.95);
\draw[dashed] (5.6,0) -- (5.6,2.95);
\node at (5.6,3.25) {median $\pm$ class tolerance};
\foreach \px/\ph in {1.2/0.5,1.5/0.7,1.8/0.4}
  {\draw[fill=black!35] (\px,0) rectangle ++(0.24,\ph);}
\node at (1.7,1.15) {occluder in front};
\foreach \px/\ph in {4.8/1.2,5.1/2.0,5.4/2.6,5.7/2.2,6.0/1.4}
  {\draw[fill=black!60] (\px,0) rectangle ++(0.24,\ph);}
\draw (6.1,2.3) -- (6.9,2.6);
\node[anchor=west] at (6.9,2.6) {object surface};
\foreach \px/\ph in {8.2/0.3,8.8/0.4,9.4/0.25,10.0/0.2}
  {\draw[fill=black!35] (\px,0) rectangle ++(0.24,\ph);}
\node at (9.2,0.95) {background leakage};
\node at (1.7,-0.8) {discarded};
\node at (5.6,-0.8) {kept};
\node at (9.2,-0.8) {discarded};
\end{tikzpicture}
\caption{The depth gate. The ranges of the points assigned to one mask
separate into an occluder in front, the surface of the object itself, and
background leaking through the mask boundary. Points outside a band around
the median range are discarded. The distribution is drawn schematically.}
\label{fig:depthgate}
\end{figure}
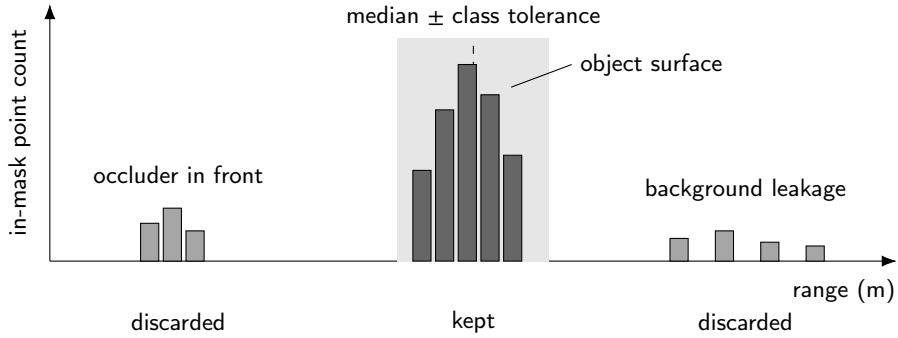

Both the depth-gate tolerance and the fit threshold are derived from the
class prior: the tolerance from the length of the class, with a floor for
the small classes, and the fit threshold at a fixed fraction of the same
length. Points on a 12-meter trailer legitimately spread over several
meters of range, while a traffic cone occupies half a meter, so a single
tolerance would either leak occluders into small objects or truncate large
ones. The range limits reproduce the class-specific ranges of the official
protocol. Barriers are not gated but segmented along their principal axis,
as described below. Figure~\ref{fig:depthgate} illustrates the gate.

The critical property of this stage is that distance is now
\emph{measured}: the position error that remains comes from which points
of the object the mask happened to capture, not from scale drift.

\subsubsection{Heuristic box fitting}

The gated points are reduced to an oriented box by deliberately simple,
class-aware, training-free rules:
\begin{itemize}
  \item \textbf{Large vehicles} (car, truck, bus, trailer, construction
  vehicle): the gated points are projected onto the ground plane and a
  rectangle is fitted to them. Yaw is searched in 30 steps of $3^{\circ}$
  over a quarter turn; for each hypothesis a rectangle is spanned by the
  2nd and 98th percentiles of the rotated points, scored by the mean
  squared distance of the points to the nearest edge, and the lowest score
  wins. Since LiDAR points concentrate on the faces visible to the sensor,
  the rectangle aligns with those faces rather than with the full object.
  The long axis is read from the observed extent when that extent exceeds
  the class threshold of Table~\ref{tab:params}, and from the viewing
  direction otherwise, since a vehicle seen head-on shows only its short
  side. Along each axis the center is the midpoint of the observed extent
  when that extent reaches the prior dimension; otherwise, when the
  observed interval lies entirely on one side of the sensor, the observed
  edge nearer the sensor is kept and the prior dimension is extended away
  from the sensor, and in the remaining case the midpoint is used. The fit
  requires at least eight gated points; below that, the center is the
  point median and the yaw is set to zero and left to the temporal layer.
  \item \textbf{Small objects} (pedestrian, bicycle, motorcycle, traffic
  cone): the point median is taken as the center; extent estimation from
  partial masks is unreliable at this scale.
  \item \textbf{Barriers}: the point set is segmented into 2.5\,m pieces
  along its principal axis. A piece is emitted only if it holds at least
  three points, and its center is the median of those points. This matches
  the annotation convention of nuScenes.
\end{itemize}
Every detection inherits the score of its source mask, and the box's
vertical position is set from the median height of the retained points.

\subsubsection{Temporal post-processing}
\label{sec:method-postproc}

A light rule-based temporal layer supplies what a single frame cannot.
Detections are first deduplicated per class by center distance, since the
same object can be segmented in two overlapping cameras or by two prompts;
the radius is larger for vehicles than for small objects, because two
pedestrians can stand closer together than two cars can park. A greedy
tracker then links detections across keyframes: each track is extended by
the nearest unclaimed detection of the same class within a radius that
grows with the number of keyframes since the track was last seen, and a
track survives a gap of at most two keyframes.

Velocity is estimated from the displacement between symmetric neighbor
keyframes, half a second apart, and is accepted only when the two
half-steps agree, either both negligible or both in the same direction.
Small magnitudes are then set to zero and large ones capped, which
suppresses spurious velocities caused by association errors. Traffic cones
and barriers are stationary by construction.

Orientation is resolved by a cascade of three rules: the direction of the
track, when the track shows genuine displacement; otherwise the velocity
direction; and for parked vehicles the driving side of the scene's city,
which settles the remaining $180^{\circ}$ ambiguity, since a parked car
almost always faces the direction of traffic on its side of the road.

\subsection{Bridging to supervised geometry}
\label{sec:method-cpbridge}

The heuristic fit of Section~\ref{sec:method-lidar} improves on camera-only
lifting, but it remains bounded by partial visibility: the mask sees one
face of the object, the LiDAR samples that face sparsely, and the fitted
center is pulled toward the visible surface. The third stage measures how
much of the remaining deficit is geometry rather than 2D detection. It
keeps the pipeline's detections, with classes and scores untouched, and
replaces only their geometry with that of a supervised LiDAR detector,
CenterPoint~\citep{centerpoint}, associated at inference time. This stage
is a diagnostic construction rather than a deployment proposal.

Association proceeds in two channels, in order, both greedy and
one-to-one: detections are processed in descending score order, and every
CenterPoint box can be claimed by at most one detection.
\begin{enumerate}
  \item \textbf{Ground-plane proximity.} Each pipeline detection claims the
  nearest same-class CenterPoint box within 2\,m in the bird's-eye view.
  \item \textbf{Image-space association.} Detections left unmatched are
  re-tried in the image. The pipeline's range errors are predominantly
  \emph{radial}, along the line of sight from the sensor, and a radial
  error barely moves a projection, so the same physical object yields
  overlapping projections even when the two boxes are meters apart in the
  ground plane. A detection claims a same-class CenterPoint box whose
  projections overlap in any camera, with a containment relaxation for the
  case where one projection sits entirely inside the other.
\end{enumerate}

\begin{figure}
\centering
\begin{tikzpicture}[font=\small]
\draw (0,0) rectangle (3.4,4.2);
\fill (1.7,0.5) circle (2.5pt);
\node at (1.7,-0.4) {sensor};
\draw[dashed] (1.7,0.5) -- (1.7,3.9);
\draw[line width=1.1pt] (1.25,1.9) rectangle (2.15,2.4);
\draw[line width=1.1pt, dash pattern=on 3pt off 2pt] (1.25,3.0) rectangle (2.15,3.5);
\draw[{Latex[length=1.6mm]}-{Latex[length=1.6mm]}] (2.45,2.15) -- (2.45,3.25);
\node[anchor=west] at (2.55,2.7) {meters apart};
\draw (5.2,1.0) rectangle (10.4,4.2);
\draw[line width=1.1pt] (6.6,1.9) rectangle (8.7,3.3);
\draw[line width=1.1pt, dash pattern=on 3pt off 2pt] (6.75,2.0) rectangle (8.8,3.25);
\node at (7.7,0.5) {projections overlap};
\node at (1.7,-1.0) {(a)};
\node at (7.8,-1.0) {(b)};
\draw[line width=1.1pt] (0,-2.0) rectangle (0.4,-1.75);
\node[anchor=west] at (0.55,-1.87) {supervised detector's box};
\draw[line width=1.1pt, dash pattern=on 3pt off 2pt] (5.6,-2.0) rectangle (6.0,-1.75);
\node[anchor=west] at (6.15,-1.87) {pipeline's box};
\end{tikzpicture}
\caption{Radial error and image-space association. (a) In the ground
plane, seen from above, two boxes for the same object sit meters apart
along the line of sight from the sensor. (b) In the camera, their
projections still overlap. The solid box is the supervised one and the
dashed box is the pipeline's.}
\label{fig:radial}
\end{figure}
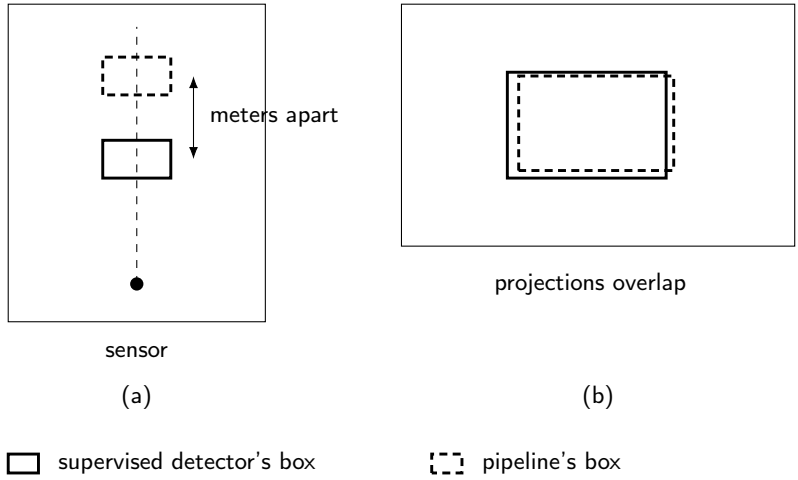

Neither threshold is fitted. The 2\,m radius is one of the four matching
distances of the official evaluation protocol, and the image-space channel
accepts an overlap of at least 0.25 intersection-over-union.
Figure~\ref{fig:radial} shows why the second channel works: the two boxes
of the same object lie meters apart along the viewing ray, yet their
projections in the camera almost coincide.

Matched detections adopt the CenterPoint box together with its orientation
and velocity; unmatched detections keep their heuristic geometry. The
second channel raises the matched share precisely on the distant objects
whose heuristic fit is worst.

\subsection{A training-free camera witness in the reverse direction}
\label{sec:method-witness}

\begin{figure}
\centering
\begin{tikzpicture}[
  font=\small,
  proc/.style={draw, rounded corners=2pt, align=center, inner sep=5pt,
               minimum height=8mm, minimum width=44mm},
  dat/.style={draw, dashed, rounded corners=2pt, align=center, inner sep=5pt,
              minimum height=8mm, minimum width=44mm},
  ar/.style={-{Latex[length=2mm]}, thick}
]
\node[dat] (cp) at (-3.4,0) {CenterPoint detections};
\node[dat, minimum width=34mm] (mk) at (3.4,0) {SAM3 masks};
\node[proc] (proj) at (-3.4,-1.5) {Project into the cameras};
\node[proc] (test) at (0,-3.0) {Overlap test against the masks};
\node[proc, minimum width=36mm] (c1) at (-4.6,-4.7) {Confirmed\\{\footnotesize same-class mask}};
\node[proc, minimum width=36mm] (c2) at (0,-4.7)    {Contradicted\\{\footnotesize other class only}};
\node[proc, minimum width=36mm] (c3) at (4.6,-4.7)  {Unwitnessed\\{\footnotesize no mask overlaps}};
\node (m1) at (-4.6,-6.1) {score $\times$ 1.00};
\node (m2) at (0,-6.1)    {score $\times$ 0.85};
\node (m3) at (4.6,-6.1)  {score $\times$ 0.60};
\node[dat, minimum width=40mm] (out) at (0,-7.5) {Re-scored detections};
\draw[ar] (cp) -- (proj);
\draw[ar] (proj) -- (test);
\draw[ar] (mk) -- (test);
\draw[ar] ([xshift=-12mm]test.south) -- (c1.north);
\draw[ar] (test.south) -- (c2.north);
\draw[ar] ([xshift=12mm]test.south) -- (c3.north);
\draw[ar] (c1) -- (m1);
\draw[ar] (c2) -- (m2);
\draw[ar] (c3) -- (m3);
\draw[ar] (m1.south) -- (out.west);
\draw[ar] (m2.south) -- (out.north);
\draw[ar] (m3.south) -- (out.east);
\end{tikzpicture}
\caption{The camera-witness rule. A supervised box is projected into the
cameras and tested against the masks of the same keyframe. Its score is
kept when a mask of the same class overlaps it, reduced when the
overlapping masks all carry a different class, and reduced further when no
mask overlaps at all.}
\label{fig:witness}
\end{figure}
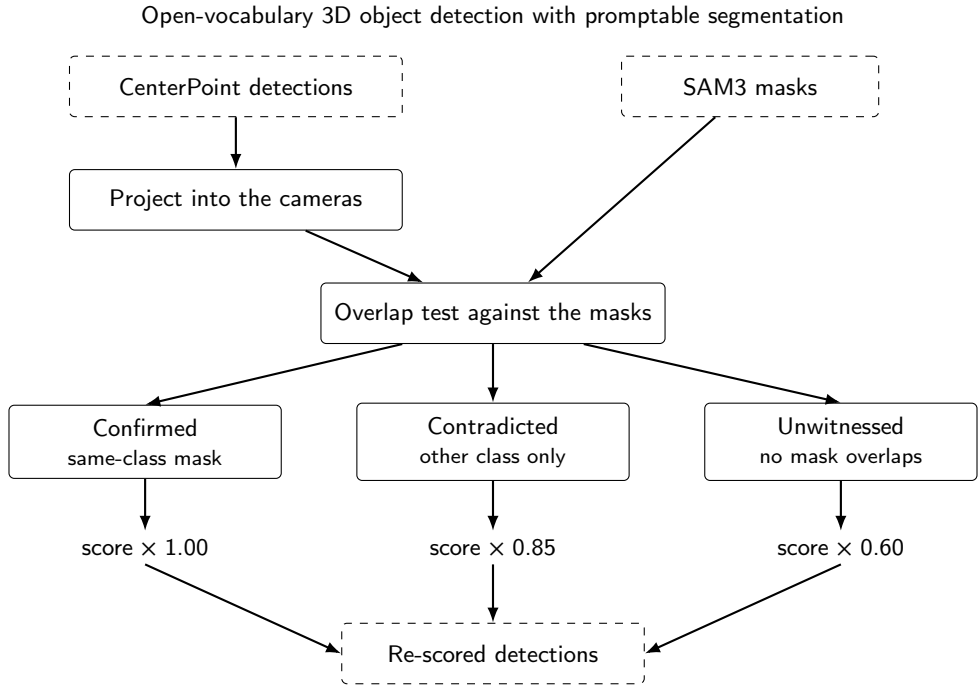

The final stage reverses the roles. CenterPoint is a LiDAR-only detector,
and the same open-vocabulary machinery is attached to it as a training-free
camera channel. Every CenterPoint box is projected into the cameras and
tested against the SAM3 masks of the same keyframe, using an image overlap
of at least 0.25 intersection-over-union; three variants of this test
differ by at most 0.003 mAP (Section~\ref{sec:witness-results}). The
outcome is one of three evidence states, each mapped to a fixed score
multiplier. A box is \emph{confirmed} when a same-class mask overlaps its
projection; its score is kept. It is \emph{contradicted} when masks
overlap but all carry a different class; its score is multiplied by 0.85.
It is \emph{unwitnessed} when no mask overlaps; its score is multiplied by
0.6. The intuition is that of a second witness: a real object should
normally be visible to the cameras, so a box that no mask corroborates is
more often a phantom, while a box the masks contradict may well have the
right geometry with only the name in dispute. No box is moved, resized or
removed; only scores change. The rule is deliberately minimal, with three
states, two hand-set multipliers that are not the best of the six settings
evaluated in Section~\ref{sec:witness-results}, and no training, so any
gain is attributable to the open-vocabulary evidence itself rather than to
a fitted model. Figure~\ref{fig:witness} summarizes the rule.

Section~\ref{sec:witness-results} compares this channel against
MVP~\citep{mvp}, which learns its camera fusion from full supervision.

\subsection{Parameters}
\label{sec:method-params}

The numeric values behind the design choices of this section are collected
in Table~\ref{tab:allparams}, and the class-dependent ones in
Table~\ref{tab:params}. The last column of Table~\ref{tab:allparams} says
where each value comes from. \emph{Protocol} and \emph{dataset} mark values
fixed by the official evaluation protocol or by nuScenes itself;
\emph{swept} marks values measured over a range, and \emph{varied} marks
the witness overlap test, for which three forms are compared. Everything
else is an engineering default, held fixed across every experiment in this
paper. The class priors of Table~\ref{tab:params} are the mean box
dimensions of each class on the training split of nuScenes, not on the
validation split: every value agrees with the training mean to within
0.2\,m, and eight of the ten to within 0.05\,m.

\begin{table}[!htbp]
\centering
\small
\setlength{\tabcolsep}{4pt}
\caption{Parameters of the pipeline. None of these values is fitted on the
evaluation split, and the last column says where each one comes from. The
class-dependent quantities are listed separately in
Table~\ref{tab:params}.}
\label{tab:allparams}
\begin{tabular}{p{56mm}p{22mm}p{50mm}p{18mm}}
\hline
Parameter & Stage & Value & Basis \\
\hline
Mask score threshold for temporal evidence & Temporal layer & 0.2 & Default \\
Minimum gated points per mask & LiDAR lifting & 3; 8 for the rectangle fit & Default \\
Barrier segment length & LiDAR lifting & 2.5\,m & Dataset \\
Duplicate-removal radius & Temporal layer & 2\,m for vehicles, 1\,m for barriers, 0.5\,m otherwise & Default \\
Track association radius & Temporal layer & 10\,m per keyframe of gap & Default \\
Largest gap a track survives & Temporal layer & 2 keyframes & Default \\
Keyframe spacing & Temporal layer & 0.5\,s & Dataset \\
Neighbor distance accepted for velocity & Temporal layer & 10\,m & Default \\
Velocity clamp & Temporal layer & below 1\,m/s set to zero, above 15\,m/s capped & Default \\
Track displacement before orientation is taken from it & Temporal layer & 3\,m and 1\,m/s & Default \\
Ground-plane association radius & Bridge & 2\,m & Protocol \\
Image-space association threshold & Bridge & 0.25 IoU & Default \\
Witness overlap threshold & Witness & 0.25 IoU & Varied \\
Witness score multipliers & Witness & 0.85 contradicted, 0.60 unwitnessed & Swept \\
\hline
\end{tabular}
\end{table}

\section{Results}
\label{sec:experiments}

\subsection{Experimental setup}

All experiments are conducted on the official nuScenes validation split of
150 scenes (6{,}019 keyframes) and evaluated with the official
\emph{DetectionEval} protocol of Section~\ref{sec:related-nuscenes}: mAP
over four center-distance thresholds, the five true-positive error
measures, and NDS as the combined score, with class-specific range limits
and ground-truth filters applied exactly as in the official
implementation. After filtering, the split contains 121{,}861 ground-truth
objects, dominated by cars (53{,}043) and pedestrians (23{,}500). The
protocol scores exactly these ten classes, so the open-vocabulary ability
of the pipeline is measured only on this fixed list, and detections
outside it count as false positives.

The compared systems are the three stages of Section~\ref{sec:method},
together with the two supervised detectors that serve as references:
\begin{itemize}
  \item \textbf{SAM3 + DVGT-2:} open-vocabulary masks lifted to metric 3D
  with the Driving Visual Geometry Transformer
  (Section~\ref{sec:method-dvgt});
  \item \textbf{SAM3 + LiDAR:} the training-free pipeline of
  Section~\ref{sec:method-lidar};
  \item \textbf{SAM3 + CenterPoint geometry:} the same detections with
  supervised box geometry associated at inference time
  (Section~\ref{sec:method-cpbridge});
  \item \textbf{CenterPoint + SAM3 witness:} the supervised LiDAR detector
  re-scored by the camera witness rule (Section~\ref{sec:method-witness});
  \item \textbf{CenterPoint} and \textbf{MVP}: the supervised reference
  detectors, LiDAR-only and LiDAR+camera respectively.
\end{itemize}
\FloatBarrier

Detection in the images uses the SAM3 image model with the official
checkpoint. Each 1600$\times$900 camera image is resized to
1008$\times$1008 for the network, and the predicted masks are mapped back
to the original size. Masks are binarized at a probability of 0.5, and the
model's own detection threshold is 0.05. Inference runs in bfloat16 on a
single NVIDIA A100 GPU, six camera images at a time. The validation split is processed once
and the resulting masks, boxes and in-mask point indices are stored, so
that every later experiment reuses the same 2D detections. Camera-only
geometry uses DVGT-2 with its official checkpoint and DINOv3 image
backbone in causal streaming mode, one frame at a time, with the trajectory
head disabled. The supervised references are the publicly released
validation predictions of CenterPoint and MVP; their scores in
Table~\ref{tab:main-results} reproduce the published validation figures,
and neither detector is retrained or given access to the pipeline's
outputs. Table~\ref{tab:versions} pins every component to the revision
that produced the results of this section.

\begin{table}[!htbp]
\centering
\small
\caption{Model and reference versions. Weights are identified by their
Hugging Face repository and code by its GitHub repository, and a version is
the first seven characters of the commit identifier. The two supervised
references are identified by the configuration name of their released
model.}
\label{tab:versions}
\begin{tabular}{lll}
\hline
Component & Source & Version \\
\hline
SAM3 weights & facebook/sam3 & 3c879f3 \\
SAM3 code & facebookresearch/sam3 & 46957e4 \\
DVGT-2 weights & RainyNight/DVGT-2 & 48fcdf2 \\
DVGT-2 code & wzzheng/DVGT & 51cf3f6 \\
DINOv3 backbone & facebookresearch/dinov3 & 6876159 \\
CenterPoint predictions & tianweiy/CenterPoint & nusc\_centerpoint\_voxelnet\_0075voxel\_fix\_bn\_z \\
MVP predictions & tianweiy/MVP & nusc\_two\_stage\_base\_with\_virtual \\
\hline
\end{tabular}
\end{table}

\subsection{Main results of the three-stage comparison}
\label{sec:main-results}

Table~\ref{tab:main-results} reports the main comparison, ordered by the
source of 3D geometry.

\begin{table}[!htbp]
\centering
\small
\setlength{\tabcolsep}{4pt}
\caption{Main results on the nuScenes validation split, ordered by the
source of 3D geometry. The upper block is training-free and open-vocabulary
in its 2D detection, and the lower block lists supervised references. The
input column gives C for camera and L for LiDAR, and lower is better in the
five right-hand columns.}
\label{tab:main-results}
\begin{tabular}{lcccccccc}
\hline
Method & Input & mAP & NDS & mATE & mASE & mAOE & mAVE & mAAE \\
\hline
SAM3 + DVGT-2                & C   & 0.183 & 0.282 & 0.691 & 0.484 & 0.843 & 0.738 & 0.342 \\
SAM3 + LiDAR                 & C+L & 0.298 & 0.348 & 0.634 & 0.347 & 0.967 & 0.733 & 0.331 \\
SAM3 + CenterPoint geometry  & C+L & 0.413 & 0.555 & 0.379 & 0.266 & 0.385 & 0.297 & 0.189 \\
CenterPoint + SAM3 witness   & C+L & 0.630 & 0.685 & 0.295 & 0.255 & 0.304 & 0.257 & 0.191 \\
\hline
CenterPoint                  & L   & 0.596 & 0.668 & 0.292 & 0.255 & 0.302 & 0.259 & 0.194 \\
MVP                          & C+L & 0.670 & 0.707 & 0.289 & 0.251 & 0.281 & 0.270 & 0.189 \\
\hline
\end{tabular}
\end{table}

Each step isolates one question. \emph{Camera-only to LiDAR} ($0.183
\rightarrow 0.298$ mAP): measured range is worth $+0.115$ mAP over learned
monocular range under an otherwise identical pipeline, the value of the
sensor separated from the value of open-vocabulary detection.
\emph{Heuristic to supervised geometry} ($0.298 \rightarrow 0.413$ mAP,
$0.348 \rightarrow 0.555$ NDS): keeping the pipeline's detections and
adopting CenterPoint boxes where a match exists lifts NDS by $+0.21$. The
largest deficit of the training-free pipeline therefore lies in
measurement precision, in position, extent, orientation and velocity,
rather than in 2D detection; Section~\ref{sec:decomposition} shows that it
is not the only one. \emph{LiDAR-only to LiDAR+camera} (lower block and
fourth row): a camera channel helps a LiDAR detector in both regimes. MVP
learns it from full supervision and gains $+0.074$ mAP over CenterPoint;
the training-free witness rule gains $+0.034$ mAP, roughly half of the
trained-fusion gain.

The five right-hand columns of Table~\ref{tab:main-results} break the
true-positive errors down across the same configurations: mATE is a
distance in meters, mAVE a speed in meters per second, mAOE an angle in
radians, and mASE and mAAE are dimensionless. Replacing the heuristic fit
by supervised boxes shrinks every term at once: translation falls from
0.634 to 0.379, orientation from 0.967 to 0.385 and velocity from 0.733 to
0.297. Scale improves least, from 0.347 to 0.266, as expected for a
pipeline that takes its extents from a class prior. What remains between
the third row and CenterPoint itself is almost entirely translation, 0.379
against 0.292. The witness rule moves no box, so its five terms are those
of CenterPoint to within 0.003, and its gain appears in mAP and NDS alone.

Table~\ref{tab:main-perclass} gives the per-class breakdown of the
training-free pipeline.

\begin{table}[!htbp]
\centering
\caption{Per-class results of the training-free pipeline on the validation
split. The true-positive error measures of the same configuration are given
in Table~\ref{tab:main-results}.}
\label{tab:main-perclass}
\begin{tabular}{lc@{\hspace{10mm}}lc}
\hline
Class & AP & Class & AP \\
\hline
Car            & 0.371 & Pedestrian   & 0.637 \\
Truck          & 0.220 & Motorcycle   & 0.454 \\
Bus            & 0.321 & Bicycle      & 0.240 \\
Trailer        & 0.009 & Traffic cone & 0.382 \\
Constr.\ veh.  & 0.114 & Barrier      & 0.233 \\
\hline
\end{tabular}
\end{table}

\subsection{The temporal layer}
\label{sec:temporal-ablation}

Table~\ref{tab:temporal} isolates the temporal layer of
Section~\ref{sec:method-postproc}. Every row uses the same detections and
the same boxes, so mAP stays at 0.298, mATE at 0.634 and mASE at 0.347
throughout, and the layer's entire contribution appears in the three
remaining error terms and in NDS. Removing it costs 0.045 NDS, and the
path between the two ends is not monotone.

\begin{table}[!htbp]
\centering
\caption{Effect of the temporal layer. Every row uses the same detections
and the same boxes, so mAP, mATE and mASE are identical throughout and only
the three remaining error terms move. The three indented rows each modify
the fifth, and the eighth is the configuration used throughout this paper.
The final row removes the velocity clamp from that configuration.}
\label{tab:temporal}
\begin{tabular}{lcccc}
\hline
Configuration & mAOE & mAVE & mAAE & NDS \\
\hline
No temporal layer, single frame               & 1.311 & 1.318 & 0.482 & 0.303 \\
Velocity                                      & 1.311 & 1.081 & 0.403 & 0.311 \\
Velocity and orientation                      & 1.099 & 1.081 & 0.403 & 0.311 \\
Narrow velocity radius with parked prior      & 1.200 & 1.259 & 0.452 & 0.306 \\
Wide velocity radius, selective parked prior  & 1.006 & 0.733 & 0.331 & 0.345 \\
\quad with the parked prior on two-wheelers   & 1.018 & 0.733 & 0.331 & 0.345 \\
\quad with velocity from the track            & 0.978 & 0.963 & 0.341 & 0.323 \\
\quad with orientation from the track         & 0.967 & 0.733 & 0.331 & 0.348 \\
Final row without the velocity clamp          & 0.967 & 0.758 & 0.333 & 0.345 \\
\hline
\end{tabular}
\end{table}

A property of the metric explains the shape of the table: NDS credits a
true-positive error term only when it falls below 1.0, so resolving
orientation in the third row improves mAOE from 1.311 to 1.099 without
moving NDS, and the gain is collected only in the fifth row, where the
velocity channel crosses below 1.0. Two rows are informative failures.
Narrowing the velocity radius zeroes the velocity of genuinely moving
objects and drags the attribute error with it. Taking velocity from the
track is worse than taking it from symmetric neighbor keyframes, because a
track that jumps identity carries that jump into its velocity; taking only
the orientation from the track is safe, and the eighth row combines the
two. The velocity clamp is the smallest component, worth 0.003 NDS.

\subsection{Sensitivity to the score threshold}
\label{sec:threshold-sensitivity}

The pipeline emits every detection whose mask score clears a low threshold
and leaves the ranking to the evaluation. Table~\ref{tab:threshold} sweeps
that threshold, with a separate official evaluation at each value, to test
whether the low value was chosen because it happens to be favorable.

\begin{table}[!htbp]
\centering
\caption{Sensitivity to the score threshold. Each row is a separate
evaluation of the same detections under the official protocol, keeping only
those above the stated score. The result changes little across the range,
so the value used is not a peak.}
\label{tab:threshold}
\begin{tabular}{cccccc}
\hline
Score threshold & mAP & mATE & mAOE & mAVE & NDS \\
\hline
0.05 & 0.298 & 0.634 & 0.967 & 0.733 & 0.348 \\
0.10 & 0.295 & 0.628 & 0.957 & 0.713 & 0.350 \\
0.15 & 0.290 & 0.625 & 0.954 & 0.701 & 0.350 \\
0.20 & 0.285 & 0.624 & 0.952 & 0.695 & 0.348 \\
0.25 & 0.280 & 0.623 & 0.951 & 0.690 & 0.347 \\
0.30 & 0.276 & 0.622 & 0.949 & 0.687 & 0.345 \\
0.40 & 0.267 & 0.620 & 0.949 & 0.683 & 0.341 \\
0.50 & 0.254 & 0.620 & 0.948 & 0.680 & 0.335 \\
\hline
\end{tabular}
\end{table}

The result is flat rather than peaked. Raising the threshold from 0.05 to
0.50 discards six of every seven detections and costs 0.044 mAP and 0.013
NDS. Even at 0.40 the pipeline emits 63{,}622 car detections against
53{,}043 annotated cars, so the sweep stays on the permissive side
throughout. The true-positive errors improve slightly, because the
detections discarded first are the weakest. AP is computed from a ranked
list, so a long low-scoring tail is close to free: it fills the low-recall
end of the curve without displacing anything above it.

\subsection{Camera-only comparison}
\label{sec:camera-comparison}

The first two rows of Table~\ref{tab:main-results} ask what a LiDAR adds
to a camera system. Adding it is worth 0.115 mAP and 0.066 NDS, and since
the 2D detections are identical, the whole difference is in placement. A
box that names the object correctly but sits a meter and a half away
matches at the two loose distances of the protocol and fails at the two
tight ones, and the tight distances are demanding even with measured
range, since the mean translation error of the LiDAR-based pipeline is
0.634\,m. The camera-only row therefore separates the two halves of the
problem: open-vocabulary detection transfers to 3D unchanged, and metric
placement does not.

\subsection{Decoupling detection from geometry}
\label{sec:decomposition}

The third stage defines what the open-vocabulary pipeline is worth when
it no longer has to supply the geometry itself
(Table~\ref{tab:main-results}, second and third rows); only the box
geometry is exchanged.

The association channel matters. Matching on ground-plane proximity alone,
within the 2\,m radius, reaches mAP~0.351 / NDS~0.519: the pipeline's
position errors at range routinely exceed 2\,m, so exactly the detections
that need supervised geometry most fail to receive it. Adding the
image-space channel of Section~\ref{sec:method-cpbridge}, which is immune
to radial error, raises the matched share from 16\,\% to 22\,\% of
detections and the result to 0.413/0.555. Two controls complete the
picture. First, the remaining gap to CenterPoint is not a matching
artifact: relabeling CenterPoint's own boxes through 2D mask association
plateaus at 0.37--0.40 mAP regardless of box quality, so the
open-vocabulary \emph{class} channel, dominated by truck--trailer and
cone--barrier confusions, is the binding constraint. Second, scores from
different sources cannot be mixed within one ranking: replacing scores for
only the matched subset of the pipeline's boxes \emph{degrades} mAP by
0.019, since the two score distributions are not on a common scale.

\subsection{Adding a camera channel to a LiDAR detector}
\label{sec:witness-results}

The three-state witness rule of Section~\ref{sec:method-witness} lifts
CenterPoint from 0.596 mAP / 0.668 NDS to 0.630 mAP / 0.685 NDS.

Two properties of this result stand out. First, the outcome barely depends
on how the witness test is scored: whether a box and a mask are matched by
projection overlap, by mask containment, or by a hybrid channel based on
LiDAR points, the result changes by less than $\pm 0.003$ mAP. What
carries the signal is whether a same-class mask exists at all, and that is
governed by visibility. Second, the comparison with MVP shows what
supervision is worth: learned camera fusion gains $+0.074$ mAP, the
training-free witness $+0.034$, so the witness recovers roughly half the
benefit of supervised fusion at zero annotation cost.

The two multipliers matter as little as the overlap test.
Table~\ref{tab:witnesssweep} evaluates six settings on the full split.
Every setting that penalizes weak evidence improves on plain CenterPoint,
and the spread across them is smaller than the gain itself. The values
used throughout this paper, 0.85 and 0.60, were fixed before this sweep
and are not the best of the six: penalizing more sharply, with 0.70 and
0.40, reaches 0.635 mAP and 0.687 NDS.

\begin{table}[!htbp]
\centering
\caption{Witness score multipliers. Each row is a complete evaluation of
the re-scored CenterPoint on the full validation split, and the row 1.00 /
0.85 / 0.60 is the setting used in this paper.}
\label{tab:witnesssweep}
\begin{tabular}{ccccc}
\hline
Confirmed & Contradicted & Unwitnessed & mAP & NDS \\
\hline
1.00 & 1.00 & 1.00 & 0.596 & 0.668 \\
1.00 & 1.00 & 0.60 & 0.621 & 0.680 \\
1.00 & 0.90 & 0.75 & 0.623 & 0.681 \\
1.00 & 0.85 & 0.85 & 0.617 & 0.679 \\
1.00 & 0.85 & 0.60 & 0.630 & 0.685 \\
1.00 & 0.70 & 0.40 & 0.635 & 0.687 \\
\hline
\end{tabular}
\end{table}

\section{Discussion}
\label{sec:analysis}

This section dissects where the training-free pipeline succeeds and where
its losses originate: what SAM3 detects in the image, how detection
degrades with visibility and distance, and why detecting an object is not
the same as scoring it.

\subsection{2D detection coverage by class}
\label{sec:analysis-2dcoverage}

The first analysis asks whether, before any 3D reasoning, SAM3 finds the
object in the image at all. For every ground-truth object within the class
range limits, the test checks whether a mask of score at least 0.2
overlaps the object's image projection at an intersection-over-union
(IoU) of at least 0.25. The strict version accepts only a mask of the same
class, the permissive version a mask of any class
(Table~\ref{tab:coverage-class}), and the gap between the two isolates the
objects that are segmented but misnamed. The population contains
134{,}565 objects, filtered only by the range limits; the official
evaluation additionally discards annotations without any LiDAR or radar
return, which gives the 121{,}861 objects of
Section~\ref{sec:experiments}.

\begin{table}[!htbp]
\centering
\caption{SAM3 2D detection coverage of in-range ground-truth objects by
class. The same-class column requires the overlapping mask to carry the
correct name, the any-class column does not, and the misnamed column is
the difference between the two. Classes are ordered by same-class
coverage.}
\label{tab:coverage-class}
\begin{tabular}{lrccc}
\hline
Class & GT & Same class & Any class & Misnamed \\
\hline
Bus            &  2{,}085 & 93.5\,\% & 95.2\,\% &  1.7\,\% \\
Car            & 59{,}522 & 90.6\,\% & 91.2\,\% &  0.6\,\% \\
Truck          &  9{,}764 & 86.8\,\% & 92.3\,\% &  5.5\,\% \\
Barrier        & 16{,}911 & 85.2\,\% & 86.7\,\% &  1.5\,\% \\
Trailer        &  2{,}446 & 79.5\,\% & 89.2\,\% &  9.7\,\% \\
Pedestrian     & 25{,}678 & 77.1\,\% & 79.5\,\% &  2.4\,\% \\
Traffic cone   & 12{,}491 & 72.1\,\% & 85.8\,\% & 13.7\,\% \\
Bicycle        &  2{,}060 & 72.1\,\% & 79.2\,\% &  7.1\,\% \\
Motorcycle     &  2{,}009 & 70.1\,\% & 82.8\,\% & 12.7\,\% \\
Constr.\ veh.  &  1{,}599 & 57.5\,\% & 76.7\,\% & 19.2\,\% \\
\hline
Total          & 134{,}565 & 84.2\,\% & 87.5\,\% & 3.3\,\% \\
\hline
\end{tabular}
\end{table}

Zero-shot coverage is high where objects are large and visually
distinctive (bus 93.5\,\%, car 90.6\,\%) and degrades for small or rare
categories. More consequential, the gap between the two versions of the
test isolates \emph{misnaming}. For
construction vehicles, one object in five is segmented but carries the
wrong label, typically ``truck''; for traffic cones the dominant competing
label is ``barrier'', and for motorcycles the two-wheeler prompts compete
with each other. For these classes the limiting factor is not whether SAM3
sees the object but what it calls it---a prompt-taxonomy problem, which
the sub-class prompts of Section~\ref{sec:method-prompts} target
directly.

\subsection{Coverage by visibility and distance}
\label{sec:analysis-visibility}

Table~\ref{tab:coverage-visdist} breaks pipeline coverage down along the
two axes that dominate it: annotated visibility and range. The measurement
here is at the \emph{box level}, whether the pipeline places a box within
4\,m of the object, and therefore reflects the full chain from mask to
fitted box. Each cell reports the number of ground-truth objects together
with their coverage; the total of 121{,}963 differs marginally from the
official count of 121{,}861 in how objects with zero sensor points are
excluded.

\begin{table}[!htbp]
\centering
\caption{Box-level coverage by annotated visibility band and range. Each
cell gives the number of ground-truth objects and the fraction of them the
pipeline covers. Visibility is the annotated fraction of the object that is
visible across the cameras.}
\label{tab:coverage-visdist}
\begin{tabular}{lcccc}
\hline
Visibility & 0--15\,m & 15--30\,m & 30\,m+ & Total \\
\hline
0--40\,\%   &  3{,}537 $\cdot$ 80\,\% & 10{,}182 $\cdot$ 59\,\% &  9{,}749 $\cdot$ 33\,\% & 23{,}468 $\cdot$ 52\,\% \\
40--60\,\%  &  1{,}915 $\cdot$ 92\,\% &  5{,}783 $\cdot$ 79\,\% &  5{,}091 $\cdot$ 54\,\% & 12{,}789 $\cdot$ 71\,\% \\
60--80\,\%  &  3{,}713 $\cdot$ 96\,\% &  7{,}964 $\cdot$ 85\,\% &  6{,}154 $\cdot$ 60\,\% & 17{,}831 $\cdot$ 79\,\% \\
80--100\,\% & 24{,}135 $\cdot$ 96\,\% & 29{,}686 $\cdot$ 87\,\% & 14{,}054 $\cdot$ 68\,\% & 67{,}875 $\cdot$ 86\,\% \\
\hline
Total       & 33{,}300 $\cdot$ 94\,\% & 53{,}615 $\cdot$ 81\,\% & 35{,}048 $\cdot$ 55\,\% & 121{,}963 $\cdot$ 77\,\% \\
\hline
\end{tabular}
\end{table}

The two axes act as independent loss channels. Within every visibility
band, coverage falls with range as the LiDAR becomes sparser; within every
range band, it falls with occlusion as masks weaken and fewer points
remain visible. At the favorable corner---near and fully visible---the
pipeline covers 96\,\% of objects; at the adverse corner it covers 33\,\%.
The training-free pipeline is thus nearly complete where a camera-based
system can be expected to operate, and its misses concentrate where the
evidence itself vanishes.

\subsection{Detecting is not scoring}
\label{sec:analysis-detect-vs-score}

Comparing 2D coverage against official AP exposes a dissociation that
class averages hide (Table~\ref{tab:catch-vs-ap}).

\begin{table}[!htbp]
\centering
\caption{2D detection coverage versus official AP for selected classes.
The two columns do not move together, which separates being found in the
image from scoring in the metric.}
\label{tab:catch-vs-ap}
\begin{tabular}{lcc}
\hline
Class & 2D coverage & Official AP \\
\hline
Trailer        & 79.5\,\% & 0.009 \\
Constr.\ veh.  & 57.5\,\% & 0.114 \\
Truck          & 86.8\,\% & 0.220 \\
Car            & 90.6\,\% & 0.371 \\
Pedestrian     & 77.1\,\% & 0.637 \\
\hline
\end{tabular}
\end{table}

Pedestrians convert 77\,\% coverage into 0.637 AP; trailers convert
79.5\,\% coverage into 0.009. The trailer failure is therefore not a
detection failure: the objects are found, but two mechanisms suppress
their score. First, their class is contested: masks on trailers frequently
say ``truck'', and vice versa, so correct detections are outscored by
confidently mislabeled ones. Second, their geometry is unforgiving: on a
12\,m articulated body, a center pulled toward the visible face misses the
2--4\,m matching thresholds that a pedestrian-sized error survives. The
same two mechanisms, in milder form, explain the car--truck gap.
Supervised geometry repairs the second mechanism, and trailer AP rises
accordingly under borrowed boxes; the class channel persists across every
geometry source and caps all mask-classified variants at 0.37--0.40 mAP.

\begin{figure}
\centering
\includegraphics[width=0.8\textwidth]{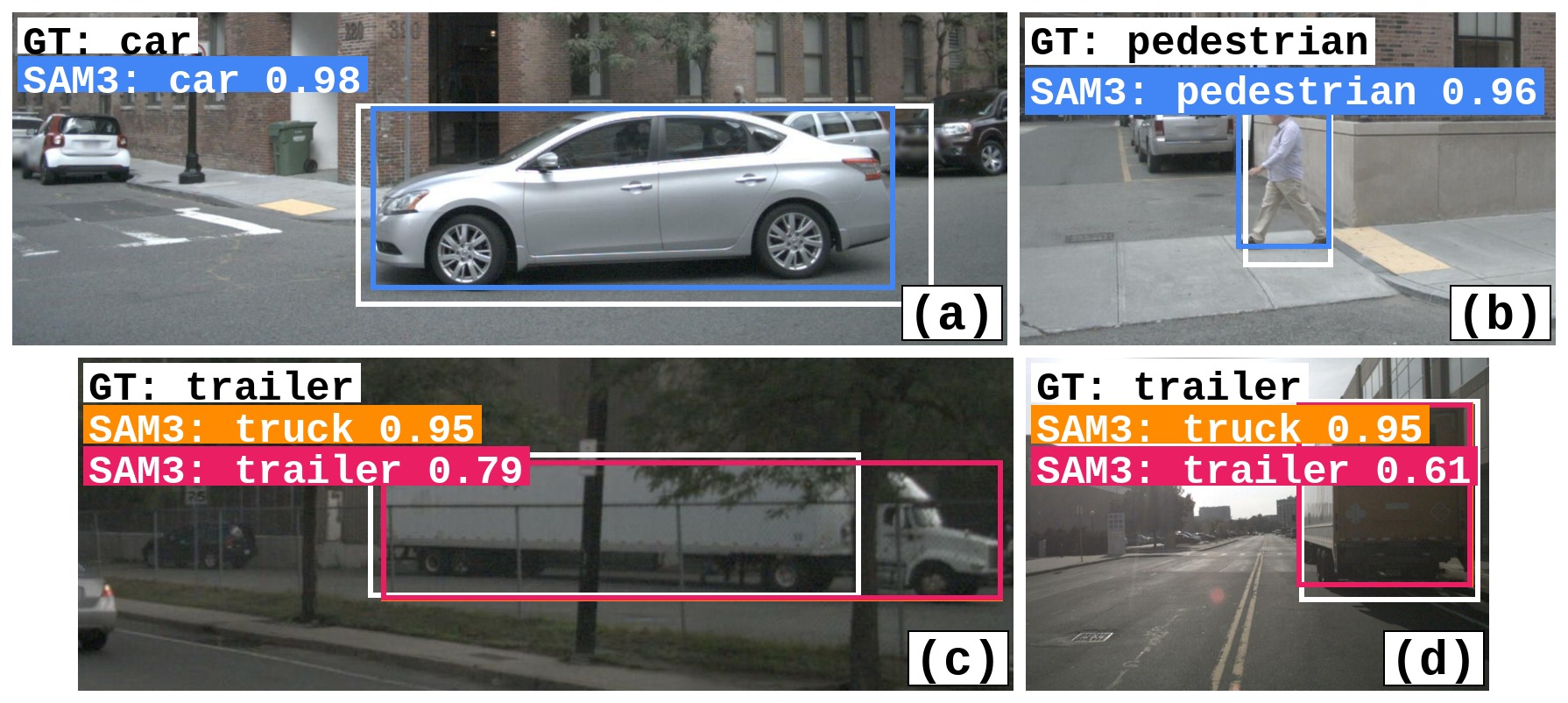}
\caption{Detection versus naming on nuScenes images~\citep{nuscenes}. (a) A
car and (b) a pedestrian carry a high-scoring mask of the correct class,
while (c) and (d) are trailers whose dominant mask says truck.}
\label{fig:misnamed}
\end{figure}

Figure~\ref{fig:misnamed} shows both sides of this dissociation. In the
top row, a car and a pedestrian each carry a high-scoring mask of the
correct class. In the bottom row, two trailers are found by SAM3, but the
dominant mask on each says ``truck'' with score 0.95, while the trailer
mask scores lower. The detection exists; its name loses the competition.

\section{Conclusion}
\label{sec:conclusion}

This paper has examined how far open-vocabulary image understanding can be
pushed into metric 3D perception without any 3D annotation, through a
three-stage comparison in which text-prompted SAM3 segmentation is fixed
and only the source of 3D geometry changes. Camera-only lifting through a
visual geometry transformer establishes that open-vocabulary detection
survives the transfer to 3D, but learned monocular range does not place
objects precisely enough for the strict center-distance metric. Fitting
boxes from the raw LiDAR points inside the same masks, with a depth gate
and heuristic rules, reaches mAP~0.298~/~NDS~0.348 under the official
nuScenes protocol at zero labeling cost. Associating supervised box
geometry to the same detections at inference time, in the ground plane
and, critically, in image space where radial error is invisible, lifts the
result to 0.413~/~0.555. Measurement precision is therefore the largest
deficit that supervised geometry removes, but not the only limit. Two
others survive the substitution: class confusion, where an object is found
but carries the wrong name, and confidence calibration, where the scores
fail to rank correct detections above incorrect ones.
Section~\ref{sec:analysis} shows that the class channel persists across
every geometry source, and that an object can be found in the image yet
score almost nothing. Reversing the direction, a three-state camera-witness
rule built from the same masks lifts a supervised LiDAR-only detector from
0.596 to 0.630 mAP, roughly half the gain of fully supervised camera
fusion, with no training.

The accompanying analysis quantifies the 2D detection layer itself. SAM3
covers 84\,\% of in-range objects with a correctly named mask, and
87.5\,\% with any mask, and coverage decays along the independent axes of
occlusion and range. The classes that fail in the official metric fail not
because they are unseen, but because they are misnamed or geometrically
unforgiving: trailers are detected in 2D at 79.5\,\% yet score 0.009 AP.
Prompt design emerges as a first-class factor: sub-class prompts
measurably improve difficult classes, and prompt wording controls what
counts as an object.

One limitation frames every number reported here. The 2D detection
component accepts free-text prompts, but the quantitative evaluation is
confined to the ten classes of the nuScenes benchmark, so the
open-vocabulary claim is measured directly only on those classes. Its
wider reach is shown indirectly, by the effect of prompt wording and by
detections that are visually correct yet fall outside the evaluated list.

Four directions follow from the analysis. The first is a modular interface
between detection and geometry: detection emits a mask, a class name and a
score, geometry consumes the mask and returns a box, and this paper shows
that three very different back-ends fit behind the same interface.
Treating both sides as replaceable modules would turn the comparison into
a general evaluation framework for new segmentation models and geometry
sources. The second is accumulating LiDAR sweeps. Each box is currently
fitted from a single sweep, and sparse returns are the main coverage
limit; deciding the existence of an object on several accumulated sweeps,
while fitting its geometry on the keyframe sweep alone, is a promising
direction. The third is class arbitration from geometry. The dominant
confusions (truck--trailer, cone--barrier, truck--construction) are
systematic, and the associated box carries strong class evidence: a 12\,m
box is not a car. Arbitrating the mask-assigned class against the matched
box is a training-free path to the class channel that currently caps
mask-classified variants. The fourth is moving beyond the fixed class
list. Visually correct detections on police vehicles count as false
positives here, because the class is excluded from evaluation; prompts
that closed-set detectors cannot express, such as \emph{police car,
ambulance, fire truck, child, stroller} and \emph{road debris}, are
exactly where open-vocabulary perception should pay off in driving, and
evaluating them requires benchmarks that annotate beyond the standard ten
classes.

\printcredits

\section*{Declaration of competing interest}
The authors declare that they have no known competing financial interests
or personal relationships that could have appeared to influence the work
reported in this paper.

\section*{Data availability}
The nuScenes dataset is publicly available. Code will be made available on
request.

\section*{Funding}
This research did not receive any specific grant from funding agencies in
the public, commercial, or not-for-profit sectors.

\section*{Declaration of generative AI and AI-assisted technologies in the
manuscript preparation process}
During the preparation of this work the authors used Claude (Anthropic) to
assist with drafting and editing the manuscript text and with typesetting
the tables and schematic diagrams from the authors' experimental results. After using this
tool, the authors reviewed and edited the content as needed and take full
responsibility for the content of the published article.

\bibliographystyle{cas-model2-names}
\bibliography{refs}

\end{document}